\documentclass[preprint,12pt]{elsarticle}

\usepackage{graphicx}
\usepackage{amssymb}
\usepackage{amsthm}
\usepackage{mathtools}
\usepackage{amsmath}
\usepackage{lineno}
\usepackage{mathrsfs}
\usepackage{booktabs, caption, makecell}
\usepackage{threeparttable}
\usepackage{algorithm}
\usepackage{algcompatible}
\usepackage{lipsum}
\usepackage{appendix}
\usepackage{float}

\usepackage[colorlinks,linkcolor=blue]{hyperref}
\usepackage[nameinlink,capitalise]{cleveref}

\DeclareMathOperator*{\argmin}{arg \, min}

\journal{Elsevier}

\begin{document}

\begin{frontmatter}
\title{A Machine Learning-based Characterization Framework for Parametric Representation of Nonlinear Sloshing}
\author[label1]{Xihaier Luo\corref{cor1}}
\ead{xluo@bnl.gov}

\author[label2]{Ahsan Kareem}

\author[label2]{Liting Yu}
\author[label1]{Shinjae Yoo}

\cortext[cor1]{Corresponding author.}

\address[label1]{Computational Science Initiative, Brookhaven National Laboratory, Upton, NY 11973, United States.}
\address[label2]{NatHaz Modeling Laboratory, University of Notre Dame, Notre Dame, IN 46556, United States.}

\begin{abstract}
    The growing interest in creating a parametric representation of liquid sloshing inside a container stems from its practical applications in modern engineering systems. The resonant excitation, on the other hand, can cause unstable and nonlinear water waves, resulting in chaotic motions and non-Gaussian signals. This paper presents a novel machine learning-based framework for nonlinear liquid sloshing representation learning. The proposed method is a parametric modeling technique that is based on sequential learning and sparse regularization. The dynamics are categorized into two parts: linear evolution and nonlinear forcing. The former advances the dynamical system in time on an embedded manifold, while the latter causes divergent behaviors in temporal evolution, such as bursting and switching. The proposed framework's merit is demonstrated using an experimental dataset of liquid sloshing in a tank under horizontal excitation with a wide frequency range and various vertical slat screen settings.
\end{abstract}


\end{frontmatter}

\section{Introduction}
\label{sec1}
Sloshing liquid in a partially filled tank is a source of concern in many engineering fields \cite{ibrahim2005liquid, Faltinsen2009sloshing}. When the external excitation amplitude is large, or the excitation frequency is close to the natural sloshing frequencies, an unrestrained free liquid surface is prone to violent wave motions. Excessive water motions escalate the hydrodynamic forces acting on the tank walls, resulting in severe structural damage, for instance, an oil leak or gas explosion \cite{hatayama2008lessons}. Vertical slat screens are one of the devices developed to suppress resonantly forced sloshing. These perforated plate-based screens effectively provide efficient wave damping by promoting vortexes in water motions and dissipating energy via turbulence cascade. Understanding and mastering the fundamental principles governing liquid-plate interactions is a necessary first step toward the development of the next generation of offshore platforms, gas containers, and other naval systems \cite{rebouillat2010fluid, maleki2008sloshing, hasheminejad2012sloshing, yu2019experimental, yu2020experimental}.

Partial differential equations (PDEs) are frequently used to describe the principles that govern physical systems in the modern world. These specified PDEs, also known as governing equations, aid in the establishment of scientific fields and the rapid advancement of technology. In the context of liquid sloshing, potential flow theory was used to simulate small-amplitude sloshing motion in a regular-configuration container \cite{faltinsen1974nonlinear, molin2013experimental}. The Laplace equation was later used to analyze nonlinear sloshing problems in two/three dimensions for more turbulent scenarios. When solved using the finite element method or the boundary element method, the computed results can capture free surface sloshing motions in various liquid storage tanks \cite{faitinsen1978numerical, cho2005finite, biswal2006non}. Another possibility is to use Navier Stokes equations to solve the problem. To simulate viscous liquid sloshing in a baffled tank with possibly broken free surfaces, several turbulence models, including Reynolds averaged Navier–Stokes (RANS) and large-eddy simulation (LES), were evaluated \cite{xue2011numerical, liu2009three, liu2016comparison}. However, when large amplitude sloshing occurs traditional finite element-based methods frequently fail to provide accurate results. To address this issue, advanced numerical methods such as arbitrary Lagrangian-Eulerian elements \cite{yong2017simulation, tang2019improved} and smoothed particle hydrodynamics \cite{iglesias2004simulation, liu2010smoothed} are introduced, where the former focuses on maintaining a high-quality mesh throughout an analysis when large deformation occurs and the latter is meshless and uses a collection of pseudo-particles to represent a given body. While knowing these equations and using advanced numerical methods allow us to simulate liquid sloshing in a variety of tanks, reliable and high-fidelity PDE-governed models remain elusive in many tank and baffle settings. 

Fortunately, the unprecedented availability of data today promises a renaissance in the analysis and understanding of physical systems. This begs the question of whether we can extract governing equations from data \cite{bongard2007automated, schmidt2009distilling, brunton2016discovering}. Historically, scientists have attempted to discover the underlying governing equations by strictly adhering to first principles such as conservation laws, geometric Brownian motion assumptions, and knowledge-based deductions \cite{hughes2005isogeometric, bathe2006finite}. However, these classical approaches are limited in their capacity to establish explicit analytical governing equations and leave a large number of complex systems unexplored, as in climate science, neuroscience, and epidemiology, to name only a few examples. On the other hand, machine learning (ML) techniques have demonstrated great success in a wide range of scientific and engineering applications, paving the way for automated discovery of governing equations and characterization of dynamical systems such as liquid sloshing \cite{jordan2015machine, duraisamy2019turbulence, brunton2020machine}.

Recently, substantial emphasis has been placed on the possibility of combining machine learning with traditional numerical and experimental methods. Xie and Zhao \cite{xie2021sloshing}, for example, integrated deep reinforcement learning and expert knowledge into the current system control method for reducing tank sloshing using two active-controlled horizontal baffles. Zhang et al. \cite{zhang2021machine} used neural networks to simulate the influence of viscous dissipation on the sloshing process, adaptively adjusting the corresponding damping coefficient defined in the numerical simulator. Teja et al. \cite{teja2021identification} suggested a convolutional neural network-based approach for classifying the forms of sloshing noise generated by vehicle brakes, namely, hit and splash. Ahna et al. \cite{ahn2019database} used artificial neural networks to forecast the severity of sloshing loads under a variety of operational and environmental conditions. Nonetheless, a very limited number of studies have been reported in the literature regarding nonlinear sloshing characterization using ML techniques. In \cite{zhang2015proper}, proper orthogonal decomposition was applied to appropriate the dynamics of wave impacts in a sloshing tank. It is noted that a rigorous numerical model is still needed when violent wave motions and long-time predictions are involved. Very recently, Moya et al. \cite{moya2019learning, moya2020physically} examined the learning ability of locally linear embedding and topological data analysis in constructing a reduced-order model for describing wave motions. This perspective relies on the wealth of information from previous measurements to guide forecasting, which can be data-intensive.

We herein address limitations in current ML-based sloshing characterization methods by introducing an integrated learning framework, which is capable of accurately detecting chaotic dynamical behaviors, for instance, signal bursting and switching, hidden in the noisy experimental measurements. Specifically, a parsimonious model combining Takens's embedding theorem and sparse learning is developed. The summary of the proposed approach is to (1) embed each sloshing sequence into a higher space via a gradient-based delay embedding method, (2) perform spectral decomposition of the embedded data, and (3) build a parameterized model based on the decomposed coordinates. As a result, the nonlinear sloshing dynamics is split into a linear and a nonlinear excitation term. Like a first-order Markov model, it allows for a fast estimation and prediction of the underlying dynamics. 

The remaining of the paper is organized as follows: \cref{sec2} presents the computational procedures of the proposed characterization framework. Next, \cref{sec3} presents the experimental validation of the proposed model. Finally, \cref{sec4} summarizes the conclusions.

\section{Methodology}
\label{sec2}

This section discusses the details of the proposed framework for parametric characterization. The learning algorithm is divided into four primary parts, which are data collection, cluster analysis, delay embedding, and sequential sparse learning.

\subsection{Step 1: data collection} 
\label{sec21}

Consider a series of measurements of some physical quantities that represent the dynamics of liquid sloshing in a partially filled container. A wave probe was used in this study to measure the free-surface elevation near the left wall of a rectangular water tank \cite{yu2019experimental, yu2020experimental}. Without loss of generality, time histories of the free surface elevations can be expressed in the following form:

\begin{equation}
    \label{eq: sec2_1}
    \mathbf{x}_{\mathbf{\theta}} = [ \underbrace{ x_{t_1}, x_{t_2}, \dots, x_{t_{k}} }_{\text{past/current states}}, \underbrace{ x_{t_{k+1}}, x_{t_{k+2}}, \dots }_{\text{future states}}]_{\mathbf{\theta}}
\end{equation}

with $x_{t_i} \in \mathbb{R}$ representing the probe measurement at time $i$ and ${\mathbf{\theta}} \in \mathbb{R}^{d_\theta}$ containing the experimental parameters, for example, the number and positions of baffles. The purpose here is to characterize observed dynamics and then predict sloshing waves, i.e., predicting future free-surface elevation states by analyzing present and prior states. We conducted a variety of trials encompassing a wide range of frequencies and baffle designs in order to develop a reliable data-driven model \cite{yu2019experimental, yu2020experimental}. Based on \cref{eq: sec2_1}, these experimental data can be synthesized into a data matrix in the following manner:

\begin{equation}
\label{eq: sec2_2}
\mathcal{D}=\left[\begin{array}{cccc}
\mid & \mid & & \mid \\
\mathbf{x}_{\theta_{1}} & \mathbf{x}_{\theta_{2}} & \cdots & \mathbf{x}_{\theta_{n}} \\
\mid & \mid & & \mid
\end{array}\right]
\end{equation}

where $n$ denotes the number of experiments.

\subsection{Step 2: cluster analysis} 
\label{sec22}
The underlying dynamics of our experimental findings are then classified using cluster analysis \cite{kaufman2009finding, aghabozorgi2015time, liao2005clustering}, a powerful method for exploratory pattern analysis and data mining. The objective of the cluster analysis is to group similar sequences $\mathbf{x}_{\mathbf{\theta}_i}$ in $\mathcal{D}$ into the same group:

\begin{equation}
    \label{eq: sec2_3}
    \mathcal{D} \xrightarrow[]{\text{CFC}} \mathcal{C}_1, \mathcal{C}_2, \dots, \mathcal{C}_K
\end{equation}

where $K$ is the number of desired clusters. Approaches typically used to cluster time series data fall into three categories: shape-based, feature-based, and model-based \cite{kaufman2009finding, aghabozorgi2015time, liao2005clustering, li2017cluster, zhang2020machine, luo2021dynamic}. Among them, feature-based approaches convert the raw time-ordered sequences into a feature vector of a lower dimension \cite{kaufman2009finding, aghabozorgi2015time}. This method is efficient in analyzing high-dimensional nonlinear dynamics and, therefore, is utilized in this paper.

\begin{algorithm}[b!]
    \caption{Compressed features-based clustering}
    \label{alg: CFC}
    \begin{algorithmic}[1]
    \STATE  {\textbf{Input:} $\mathcal{D}$ and $K$}
    \STATE  {Compute the feature vector of each sequence $\mathbf{f}(\mathbf{x}_{\mathbf{\theta}_i})$}
    \STATE  {Encode the feature vector to a latent space with lower dimension by proper orthogonal decomposition: $\mathbf{f} \longmapsto \mathbf{z}$}
    \STATE  {Assign initial values for $m_1, m_2, \dots, m_K$}
    \WHILE{$\epsilon_t<\epsilon$ and $t$ is smaller than the maximum number of iterations}
    \FOR{$i=1, 2, \dots, n$}
    \STATE  {Assign each encoded vector $\mathbf{z}_i$ to the cluster which has the closest mean}
    \STATE  {Calculate new mean for each cluster}
    \ENDFOR
    \STATE  {Update iteration number $t = t + 1$}
    \ENDWHILE
    \STATE  {\textbf{Output:} clustered data $\mathcal{C}_1, \mathcal{C}_2, \dots, \mathcal{C}_K$}
    \end{algorithmic}
\end{algorithm}

The extracted features dictate the performance of feature-based techniques. It's worth noting that some features that function well on one sequence may not work well on another. While expanding the feature map by adding more features might improve its generalization capabilities, it also increases computational complexity linearly \cite{aghabozorgi2015time, liao2005clustering}. To combine computational efficiency and model performance, we present a compressed features-based clustering technique (CFC) for quick cluster analysis of embedded sloshing sequences. The proposed CFC algorithm's detailed computation procedures are detailed in \cref{alg: CFC}.

\subsection{Step 3: delay embedding} 
\label{sec23}

Many studies have demonstrated that the state of a deterministic dynamical system is the information required to faithfully characterize the evolution of the system \cite{takens1981detecting}. In this context, it means a free-surface elevation sequence recorded as a vector can be used to gain information about the dynamics of the entire state space of the system. However, the intrinsic dynamics associated with each cluster are highly nonlinear, even chaotic, consisting of various complex dynamic interactions, including standing waves, broken free surfaces, overturning waves, etc \cite{ibrahim2005liquid, Faltinsen2009sloshing}. To better characterize the sloshing sequence of the interest, we consider the application of time delay embedding, i.e., structuring data prior to analysis by stacking the state variable at different time instances:

\begin{equation}
    \label{eq: sec2_4}
    \boldsymbol{\chi}_{t} = [ x_{t}, x_{t+\tau}, \dots, x_{t+\left( d-1 \right)\tau} ]^{T} \in \mathbb{R}^{d \times 1}
\end{equation}

where $\tau$ is the delay step controlling the volume of temporal separation between two adjacent elements in $\boldsymbol{\chi}_{t}$ and $d$ is the embedding dimension determining the spread-out projection $x \in \mathbb{R} \rightarrow \boldsymbol{\chi} \in \mathbb{R}^{d}$. In practice, a reasonable choice of embedding parameter $\tau$ and $d$ is critical as they both directly affect the representation accuracy of the embedded model. Many methods have been developed to select these two parameters \cite{broomhead1986extracting, kennel1992determining, kim1999nonlinear, grassberger2004measuring}. In this paper, the so-called average mutual information (AMI) \cite{kennel1992determining} is adopted to determine the delay step $\tau$ and the false nearest neighbors (FNNs) method \cite{grassberger2004measuring} is utilized to determine the embedding dimension $d$.

\subsection{Step 4: sequential sparse learning} 
\label{sec24}

With embedded vector $\boldsymbol{\chi}_{t}$ containing enriched dynamics, we can transform a time series into a matrix of time–dependent chunks of data:

\begin{equation}
\label{eq: sec2_5}
\boldsymbol{X}=\left[\begin{array}{cccc}
\mid & \mid & & \mid \\
\boldsymbol{\chi}_{1} & \boldsymbol{\chi}_{2} & \cdots & \boldsymbol{\chi}_{k} \\
\mid & \mid & & \mid
\end{array}\right]
\end{equation}

The objective is to develop a model for $\boldsymbol{X}$ that is both characterized and predictive. To accomplish this, we will investigate a recently presented model dubbed Hankel's alternative view of Koopman (HAVOK) \cite{brunton2017chaos, li2021novel}. To characterize the dynamics, the HAVOK model first factorizes aaa using spectral decomposition methods. Singular value decomposition (SVD) is used in this case because elimination-based approaches such as LU decomposition are statistically unstable when working with a nonsquare matrix \cite{golub1971singular}. The SVD technique is as follows:

\begin{equation}
    \label{eq: sec2_6}
    \boldsymbol{X} = \boldsymbol{U} \Sigma \boldsymbol{V}^{T}
\end{equation}

where left singular vectors are defined as the columns of an orthogonal matrix $\boldsymbol{U} \in \mathbb{R}^{d \times d}$ and they represent eigenvectors of $\boldsymbol{X} \boldsymbol{X}^{T}$, right singular vectors are returned as the columns of $\boldsymbol{V} \in \mathbb{R}^{n_{t} \times n_{t}}$ and they represent eigenvectors of $\boldsymbol{X}^{T} \boldsymbol{X}$, and nonnegative singular values $\sigma_1 \geqslant \sigma_2 \geqslant \dots \sigma_k \geqslant \sigma_{k+1} = \dots = \sigma_{n_t} = 0$ are entries of the diagonal matrix $\Sigma \in \mathbb{R}^{d \times n_{t}}$ with $k = \text{rank} \left( \boldsymbol{\chi} \right)$. Such normalization and orthogonality properties of singular vectors $\{ V_{i} \}_{i=1}^{r}$ allow an efficient way of modeling dynamics using a finite number of SVD modes \cite{taira2017modal, kutz2016dynamic}:

\begin{equation}
    \label{eq: sec2_8}
    \frac{d \boldsymbol{V}_{r} \left( t \right)}{dt} = \mathcal{F} \left( \boldsymbol{V}_{r} \right) = \boldsymbol{\alpha} \boldsymbol{V}
\end{equation}

with $r$ denoting the truncation number. In \cref{eq: sec2_8}, the left side term $\frac{d V_{i} \left( t \right)}{dt}, i=1, \dots, r$ can be computed by standard numerical differentiation methods such as fourth-order central finite difference method, and the right side operator $\mathcal{F} \left( \cdot \right)$ contains regression coefficients that need to be determined. A closed linear model stated in \cref{eq: sec2_8}, however detailed, may not be sufficient to capture and reproduce the nonlinear characteristics of a highly nonlinear dynamical system. In \cite{brunton2017chaos, khodkar2021data}, it is suggested to add a forcing term to the linear model to better capture the nonlinear dynamics:

\begin{equation}
    \label{eq: sec2_9}
    \frac{d \boldsymbol{V}_{r} \left( t \right)}{dt} = \boldsymbol{\alpha}^{\dagger} \boldsymbol{V}^{\dagger} + \boldsymbol{\alpha}_r V_r
\end{equation}

where $\boldsymbol{V}_r$ is split into two parts, i.e. the linear part $\boldsymbol{V}^{\dagger} = \{ V_1, \dots, V_{r-1} \}$ and the nonlinear forcing part $V_r$. The next goal is to identify regression coefficients $[\boldsymbol{\alpha}^{\dagger}, \boldsymbol{\alpha}_r]$ by sparse learning. To compute the optimal coefficients, the least-squares (LS) method is adopted. In principle, the LS method minimizes the residual sum of squares:

\begin{equation}
    \label{eq: sec2_10}
    \boldsymbol{\alpha}^{\star} = \argmin \sum_{i=1}^{n_{t}} \left( \boldsymbol{V}_r^{'} \left( t_i \right) - \sum_{j=1}^{r} V_j \boldsymbol{\alpha}_j \right)^2
\end{equation}

However, \cref{eq: sec2_10} is ill-conditioned and numerically unstable in the context of embedded sloshing dynamics for at least two reasons. First, matrix $\boldsymbol{V}_r$ is overdetermined where $n_t \gg r$, indicating \cref{eq: sec2_10} may not produce a unique solution. Secondly, the inevitable noise integrated in the experimental data intensifies the numerical round-off error when inverting $\boldsymbol{V}_r$ to evaluate coefficients $\left( \boldsymbol{V}_r^T \boldsymbol{V}_r \right)^{-1} \boldsymbol{V}_r^T \boldsymbol{V}_r^{'}$. To approximate the nonlinear dynamics more accurately, shrinkage method \cite{tibshirani1996regression} is utilized to estimate the optimal coefficients:

\begin{equation}
    \label{eq: sec2_11}
    \boldsymbol{\alpha}^{\star} = \argmin || \sum_{i=1}^{n_{t}} \left( \boldsymbol{V}_r^{'} \left( t_i \right) - \sum_{j=1}^{r} V_j \boldsymbol{\alpha}_j \right) ||^2_2 + || \lambda \sum_{j=1}^{r} \boldsymbol{\alpha}_j ||^2_2
\end{equation}

where $\lambda$ is the regularization parameter that controls the shrinkage: as $\lambda \rightarrow 0, \boldsymbol{\alpha}^{\star} \rightarrow \boldsymbol{\alpha}^{LS}$, and as $\lambda \rightarrow \infty, \boldsymbol{\alpha}^{\star} \rightarrow \boldsymbol{0}$. Often referred to as the ridge regression, \cref{eq: sec2_11} places quadratic constraints on model coefficients $\boldsymbol{\alpha}$. The analytical ridge solution of this minimization problem takes the form of:

\begin{equation}
    \label{eq: sec2_12}
    \boldsymbol{\alpha}^{\star} = \left( \boldsymbol{V}_r^T \boldsymbol{V}_r + \lambda \boldsymbol{I} \right)^{-1} \boldsymbol{V}_r^T \boldsymbol{V}_r^{'}
\end{equation}

\begin{algorithm}[b!]
    \caption{Sequential coefficient optimization}
    \label{alg: SSL}
    \begin{algorithmic} [1]
    \STATE  {\textbf{Input:} $\boldsymbol{V}_r$, $\boldsymbol{V}_r^{'}$, $\lambda$, $\epsilon$}
    \STATE  {\textbf{Begin}}
    \WHILE{$\hat{\boldsymbol{\alpha}} \neq \hat{\boldsymbol{\alpha}}^{\star}$}
    \STATE  {Initialize the set $\mathcal{C} = \{ \}$}
    \STATE  {Evaluate the coefficient vector $\hat{\boldsymbol{\alpha}}$ via \cref{eq: sec2_12}}
    \FOR{$i=1, 2, \dots, r$}
    \IF{$\hat{\boldsymbol{\alpha}} \left( i \right)< \epsilon$}
    \STATE  {$\hat{\boldsymbol{\alpha}} \left( i \right) = 0$}
    \ELSE
    \STATE  {Add iterative counter $i$ to set $\mathcal{C} = \{ \}$}
    \ENDIF
    \ENDFOR
    \STATE  {Update the right-hand side matrix $\boldsymbol{V}_r = \boldsymbol{V}_r [ :, \mathcal{C}]$ }
    \STATE  {Update the coefficient vector $\hat{\boldsymbol{\alpha}}^{\star}$ via \cref{eq: sec2_12}}
    \ENDWHILE
    \STATE  {\textbf{Output:} coefficients matrix $\hat{\boldsymbol{\alpha}}^{\star}$}
    \end{algorithmic}
\end{algorithm}

with $\boldsymbol{I}$ denoting an identity matrix of $\mathbb{R}^{r \times r}$. By adding positive quantities to the diagonal entries, the non-full-rank condition of sparse learning encountered in the LS method is partially alleviated, and the coefficients estimator becomes nonsingular \cite{tibshirani1996regression}. To further enhance the uniqueness of the optimized $\boldsymbol{\alpha}^{\star}$, the ridge regression is implemented in a sequential thresholded fashion, where columns in $\boldsymbol{V}_r$ correspond to an insignificant coefficient is automatically deleted \cite{rudy2017data}. Steps to implement this sequential optimization algorithm are summarized in \cref{alg: SSL}. 

\subsection{Review of framework} 
\label{sec25}

\begin{figure}[b!]
    \centering
    \includegraphics[width=0.90\textwidth]{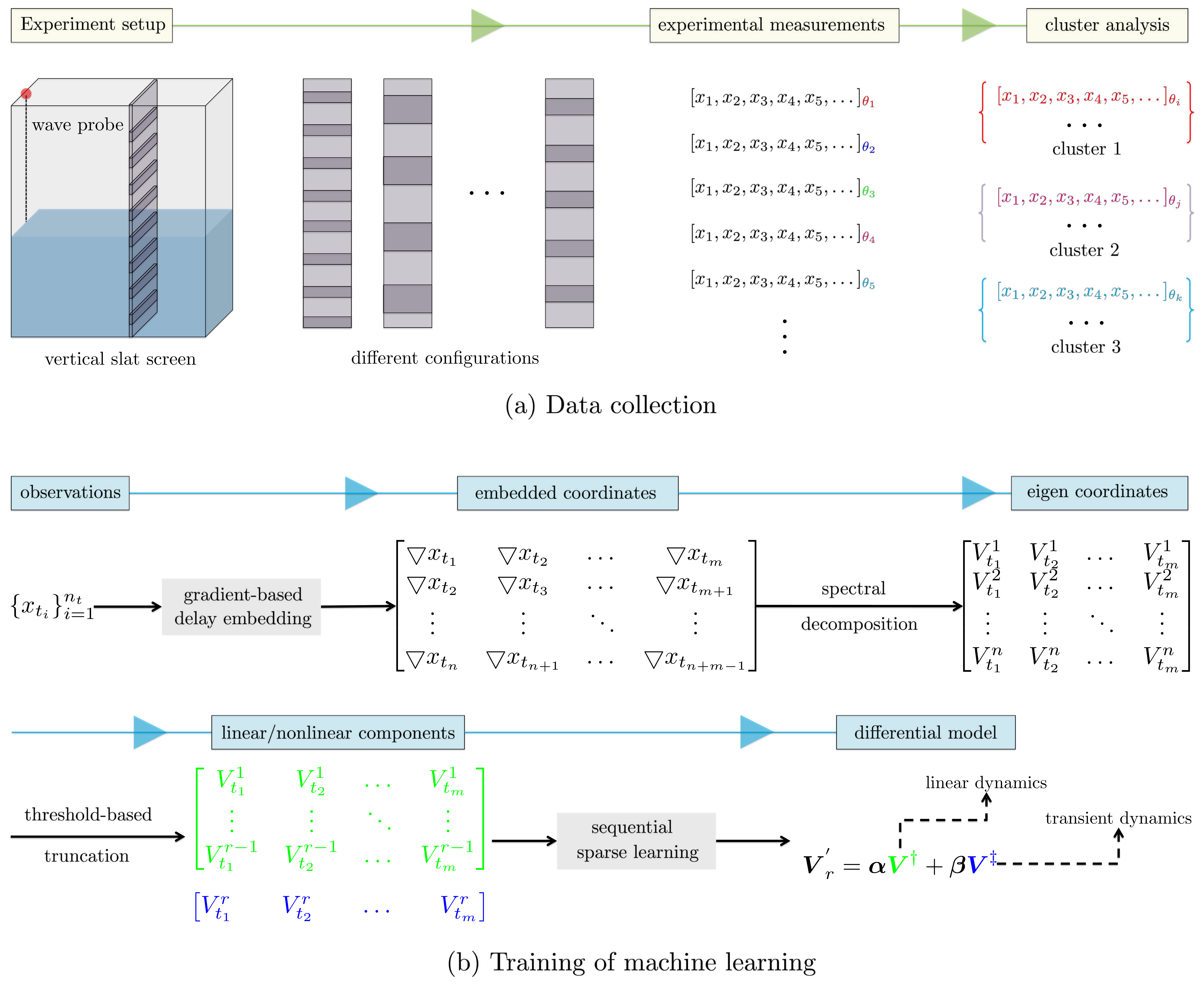}
    \caption{Overview of the machine learning model for modeling of liquid sloshing.} 
    \label{fig: f1}
\end{figure}

The preceding four steps are the integral elements of the proposed machine learning framework. \cref{fig: f1} depicts the data flow across the framework graphically. There are two advantages to the proposed characterization framework that should be discussed. First, the overall framework is very adaptable, allowing one or more elements to be replaced with other data-driven models. Different off-the-shelf SVD solvers, for example, can be simply integrated within the framework. Second, no prior knowledge of data structure or topology is required. The framework is implemented automatically, making it accessible to both domain scientists and non-experts. As a result, this framework has high computational efficiency and significant deployment flexibility.

\section{Results and Discussion}
\label{sec3}
This section analyzes the results of the machine learning. More precisely, we examine the findings of the characterization in light of the framework's four stages.

\subsection{Part 1: data collection results}
\label{sec31}
\begin{figure}[H]
    \includegraphics[width=1.0\textwidth]{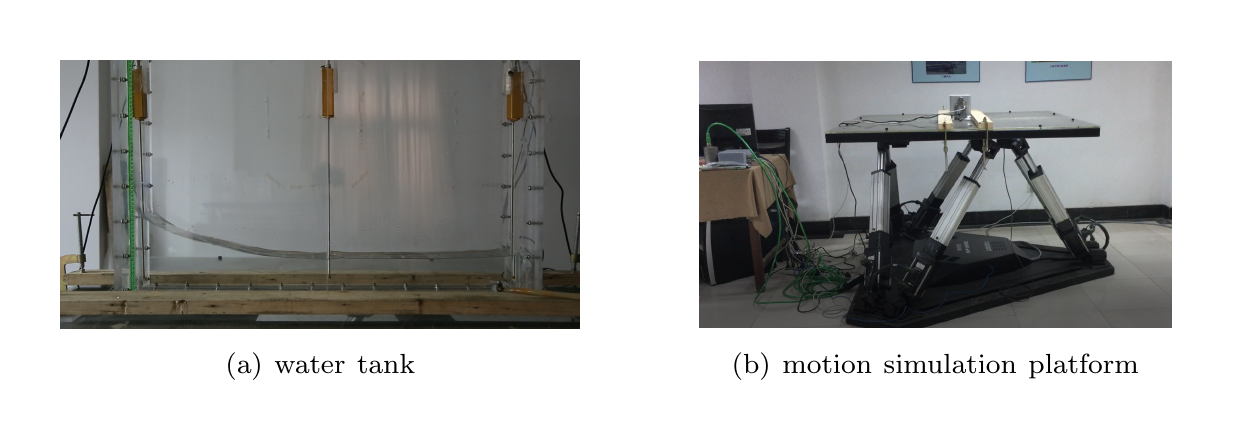}
    \caption{Experimental set-up: shake table with shallow water sloshing.} 
    \label{fig: f2}
\end{figure}

The experimental data for this work was provided by the Laboratory of Vibration Test and Liquid Sloshing at Hohai University in China \cite{yu2019experimental, yu2020experimental}. A six-degree-of-freedom motion simulation platform, also known as a hexapod, was used in particular to generate the liquid tank's motions based on a specified input of time histories (See \cref{fig: f2}). A rectangular plexiglass tank with internal dimensions of $1000 \, mm \times 700 \, mm \times 100 \, mm$ (length $\times$ height $\times$ width) is employed. It is $8 mm$ thick and partially filled with water, with the liquid depth-to-tank-length ratio set at $0.12$. Shallow water sloshing is suppressed by vertical slat screens composed of perforated aaa steel plates. Slat screens have three different solidity ratios ($0.4, 0.6$, and $0.9$) and two different slot diameters ($5 \, mm$ and $50 \, mm$), yielding six possible testing setups.

Furthermore, a horizontal sinusoidal motion defined by $X(t) = A \sin (\omega t)$ is applied to the platform. Note $A/l=0.01$ with $l$ denoting the length of the tank. The excitation frequency $\omega$ is set up by a control computer and a wide range of frequencies have been tested. The displacement of the motion platform was recorded by a displacement sensor. A wave probe (WH200) located at a distance of $15 \, mm$ from the left vertical wall is used to track the evolution of the free-surface elevations near the tank wall. As a result, a total of $539$ sequences, where each sequence corresponds to a specific experiment configuration, have been recorded for analysis. A more detailed experimental setup is available in \cite{yu2019experimental, yu2020experimental}.

\subsection{Part 2: clustering results}
\label{sec32}
\begin{figure}[H]
    \centering
    \includegraphics[width=0.97\textwidth]{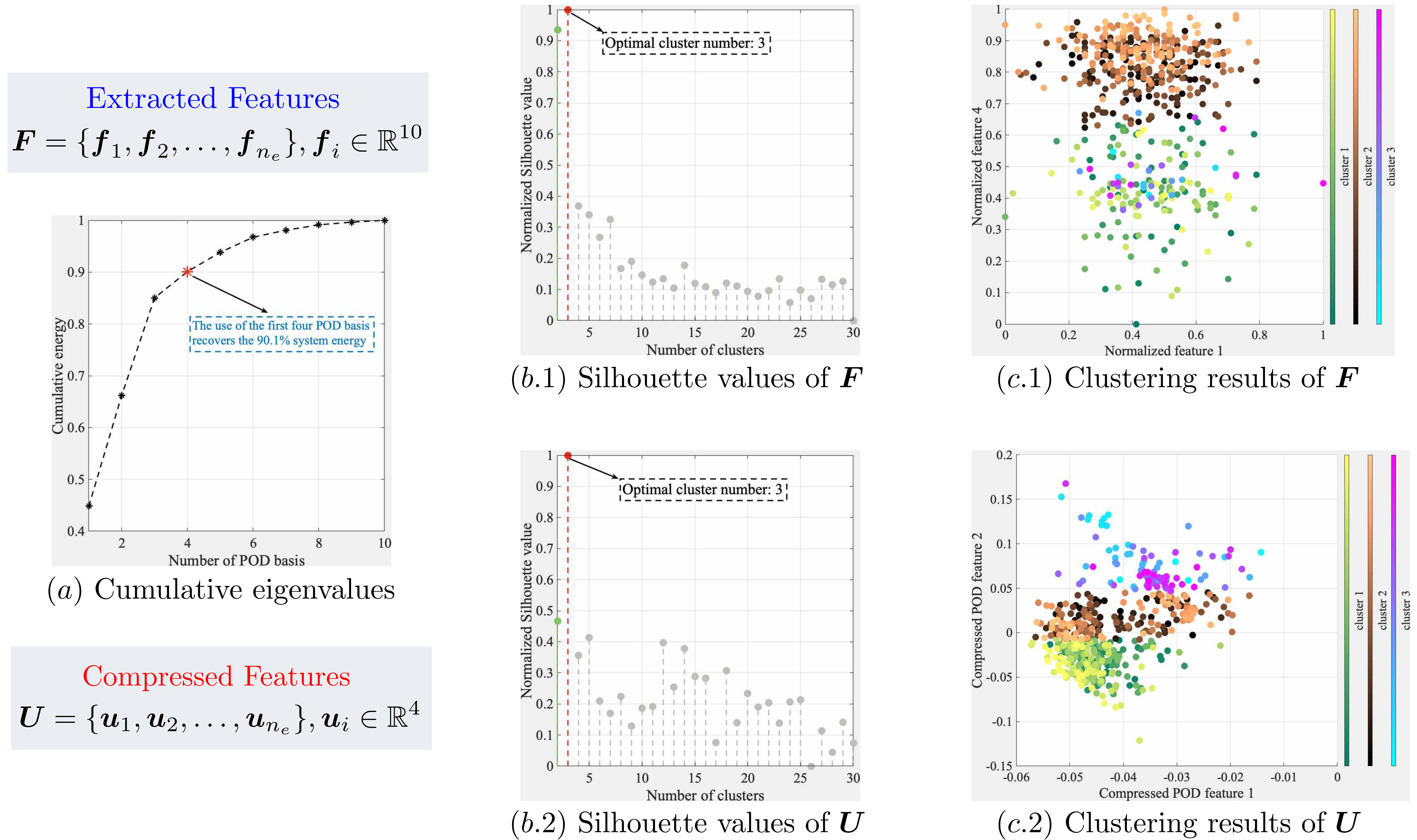}
    \caption{Clustering results: (a) The ratio of the cumulative sum of eigenvalues over the total sum of them; (b) Silhouette analysis for k-means clustering on sequence data with $n_c=2,3,\dots,30$; and (c) The visualization of the clustered data.} 
    \label{fig: f4}
\end{figure}

Direct cluster analysis is used to demonstrate the efficiency of the proposed compressed features-based clustering algorithm (See \cref{alg: CFC}). The eigenvalues of the feature matrix are initially arranged in decreasing magnitude. As a result, the estimated eigenvalue's cumulative proportion of variance can be examined. The cumulative eigenvalues are shown in \cref{fig: f4} (a). It can be seen that retaining the first four POD compressed features $\boldsymbol{u} \in \mathbb{R}^{4}$ can provide a good representation of extracted features $\boldsymbol{f} \in \mathbb{R}^{10}$ with an adequate level of accuracy, accounting for $90.1 \%$ of the variance.

Then, Silhouette analysis is used to determine the best cluster number for both approaches \cite{rousseeuw1987silhouettes}. The Silhouette value $s$, in particular, is a measure of the separation distance between the clustering results, and it has a range of $[-1, 1]$. To illustrate, the computed Silhouette values have been normalized to $[0, 1]$, with $s = 1$ indicating that the given dataset is well clustered because each sequence is clearly separated from the neighboring clusters, and $s = 0$ indicating that the select clustering algorithm fails to assign each sequence due to ambiguous boundary lines between two clusters. In both cases, the computed results presented in \cref{fig: f4} (b.1) and (b.2) imply that the optimal cluster number is $3$. In direct clustering, $n_{c} = 2$ and $n_{c} = 3$ produce similar Silhouette values, whereas $n_{c} = 3$ outperforms other values in the clustering evaluation using POD compressed features. As a result, the number of clusters is set to $3$.

Following that, the uncompressed and compressed feature matrices $\boldsymbol{F} \in \mathbb{R}^{10 \times n_e}, \boldsymbol{U} \in \mathbb{R}^{4 \times n_e}$ are partitioned into $3$ clusters. Because the quality of the initial centroid points has a direct impact on the clustering results, a widely used heuristic approach, the k-means++ algorithm, is used to improve the quality of the center initialization in this NP-Hard problem setting \cite{aghabozorgi2015time, liao2005clustering}. The maximum number of iterations for the k-means++ algorithm is $1000$, and the Minkowski distance with $p=2$ is used to compare the similarity of two different sequences. The computed results are summarized in \cref{fig: f4} (c.1) and (c.2). It can be seen that the POD feature-based clustering produces more visually distinct results. The direct method tends to mix data from the second and third clusters, whereas the proposed method can provide a clear boundary line between clusters 2 and 3.

\begin{figure}[H]
    \centering
    \includegraphics[width=0.9\textwidth]{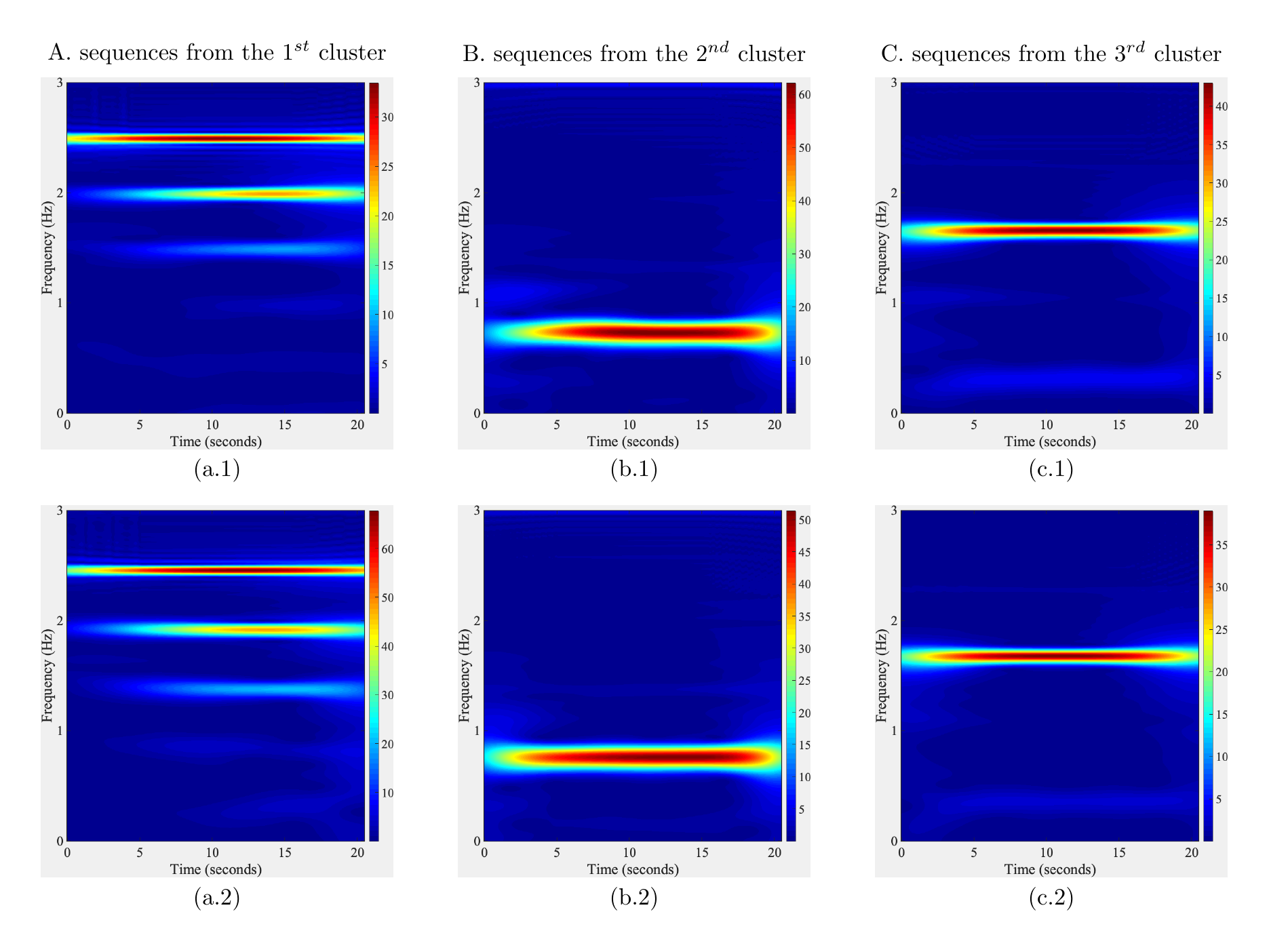}
    \caption{Wavelet analysis results of clustered free-surface elevation sequences.} 
    \label{fig: f5}
\end{figure}

Wavelet transforms are applied to the clustered sequences to further validate the clustering results. Three sequences are chosen at random from each cluster for demonstration. \cref{fig: f5} summarizes wavelet results as well as experimental parameters for the chosen sequences. Three critical properties should be discussed here. First, frequencies close to the first natural frequency are found to maximize the computed wavelet scalogram \cite{yu2019experimental}. When compared to the second and third clusters, selected sequences from the first cluster have a higher dominant frequency, that is, $\omega_{c_1} \in [2, 3]$ while $\omega_{c_2} \in [0, 1]$ and $\omega_{c_3} \in [1, 2]$. This is due to the fact that the excitation frequency used is close to the peak frequency of sloshing in the tank without baffles (It is 0.54 Hz according to the experiments). When the tank is outfitted with porous baffles, violent sloshing is reduced, the liquid is separated into zones, and the moving amplitude is suppressed \cite{yu2020experimental}. Second, several frequencies are significant for sequences chosen from the first cluster (See \cref{fig: f5} (a.1) and (a.2)). Furthermore, as sloshing evolves, the scalograms of the first cluster tend to exhibit more complicated nonlinear characteristics within the limits of the transformed resolution. This observation agrees with the experimental finding presented in \cite{yu2019experimental}, which states that when passive control devices (e.g., internal baffles) are used to suppress the maximum free-surface elevations resonated at the first natural frequency, more sloshing phenomena occur at the new frequencies. Third, the second cluster sequences have a higher excitation frequency, whereas the other two clusters have higher harmonic periods. Despite the fact that cluster 2 and cluster 3 sequences have only one dominant frequency band, the bandwidth is strongly influenced by the external excitation frequency (See \cref{fig: f5} (B) and (C)). Overall, the wavelet-based time-frequency representation provides some insight into the timing and frequencies at which a sloshing sequence intensifies. More importantly, it confirms that the proposed clustering algorithm successfully grouped sloshing sequences with similar dynamics into the same cluster.

\subsection{Part 3: embedding results}
\label{sec33}
First, the delay step $\tau$ is estimated by computing the AMI value. Different initialization parameters $\tau=\Delta t, 2 \Delta t, \dots, 20 \Delta t$ have been substituted into \cref{eq: sec2_1}. The probability distributions involved in the computation of AMI are estimated using histograms on $\boldsymbol{x}_{t}$ and $\boldsymbol{x}_{t+\tau}$, where intervals $[\text{min}(\boldsymbol{x}_{t}), \text{max}(\boldsymbol{x}_{t})]$ and $[\text{min}(\boldsymbol{x}_{t+\tau}), \text{max}(\boldsymbol{x}_{t+\tau})]$ are divided into $N_{t}$ and $N_{t+\tau}$ sub-intervals \cite{kennel1992determining}. And probabilities $p \left( x_{t}\right), p \left(x_{t+\tau} \right), \text{and } p \left( x_{t}, x_{t+\tau} \right)$ are estimated by counting the number of values within $N_{t}$, $N_{t+\tau}$, and $N_{t, t+\tau}$ sub-intervals and dividing each of these counts by $n_t = 2500$.

\begin{figure}[H]
    \centering
    \includegraphics[width=0.87\textwidth]{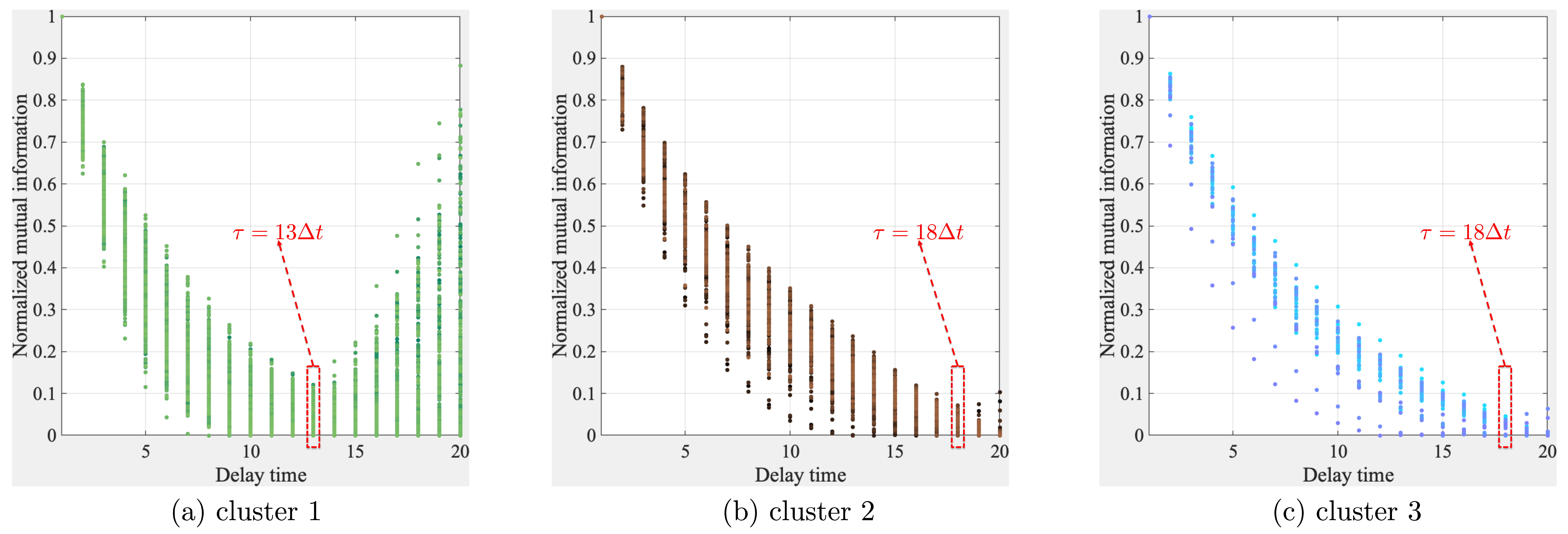}
    \caption{Selection of delay step $\tau$.} 
    \label{fig: f6}
\end{figure}

\cref{fig: f6} shows the computed results for each cluster. For each candidate $\tau$, the plot vertically displays the distribution of the computed AMI values for sequences within that cluster. The minimum AMI in an average sense occur at $\tau = 13 \Delta t, 18 \Delta t$ and $18 \Delta t$ for cluster $1, 2$ and $3$, respectively. Especially for cluster $1$, AMI value increases rapidly from $13 \Delta t$ to $20 \Delta t$. This can be explained by the dominant frequency of the first cluster. Shown in the first row of \cref{fig: f5}, sequences in this cluster have multiple controlling frequencies. As $\tau$ surpasses the intrinsic frequency, redundant information at other frequencies begins to dominate. $\boldsymbol{x}_{t}$ and $\boldsymbol{x}_{t+\tau}$ are no longer sufficiently independent, giving rise to the increases of AMI value. Therefore, the delay step is set as the candidate corresponds to the actual first minimum of AMI, namely, 13, 18 and 18.

\begin{figure}[H]
    \centering
    \includegraphics[width=0.87\textwidth]{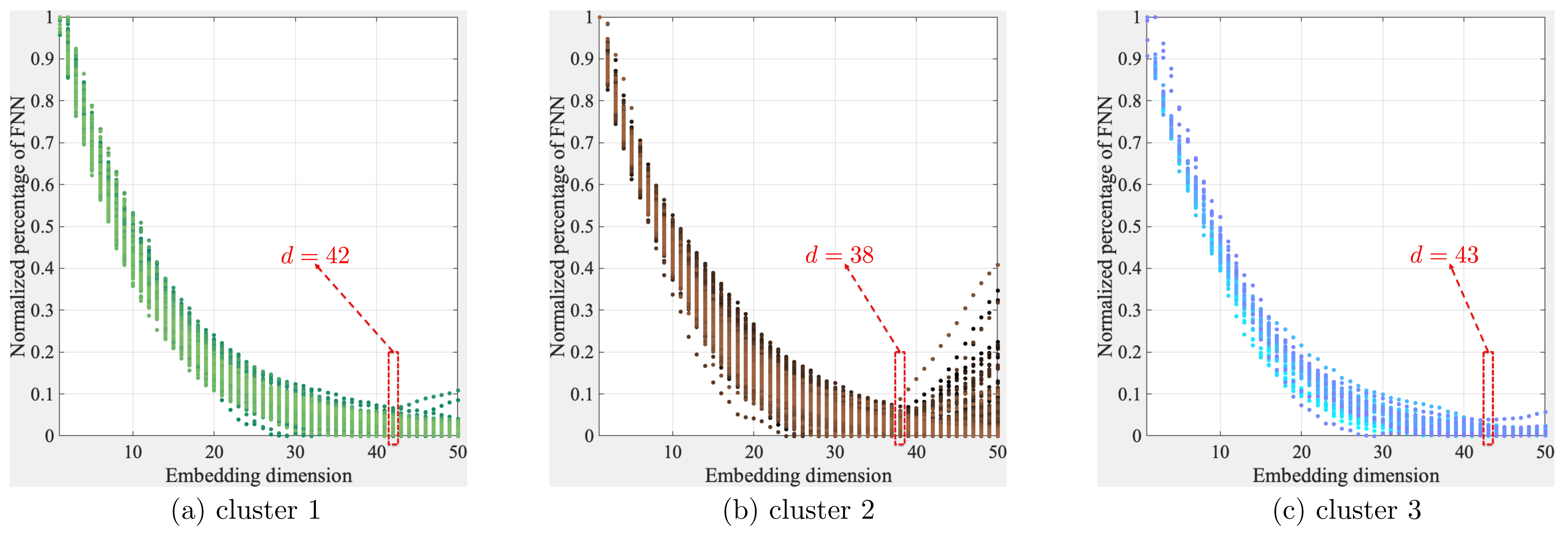}
    \caption{Selection of embedding dimension $d$.} 
    \label{fig: f7}
\end{figure}

Next, the embedding dimension can be selected by the FNNs method. The relation curve of the embedding dimension and FNNs percentage is presented in \cref{fig: f7}. Similarly, the dots associated with a specific dimension represent the distribution of the FNN percentage of different sequences in a cluster. It can be seen that these percentages drop by $90 \%$ when $d \approx 35$, $30$, and $30$ for cluster $1$, $2$, and $3$, respectively, indicating the dimension where the sloshing dynamics of interest can be smoothly unfolded \cite{}. Meanwhile, it should be stressed that the percentage of a few sequences starts to increase (e.g., the tail parts of \cref{fig: f7} (b)). This phenomenon is often referred to as the \textsl{noise to signal} effects, where experimental noises overpower the FNNs statistics \cite{kim1999nonlinear, grassberger2004measuring}. In many applications, this turning point serves as an indicator of the level of single contamination, providing a way to distinguish chaotic dynamics from noise perturbations. In the following analysis, the embedding dimension is set to $d=42$, 38 and 43 for cluster $1$, $2$, and $3$, respectively.

\begin{figure}[H]
    \centering
    \includegraphics[width=0.93\textwidth]{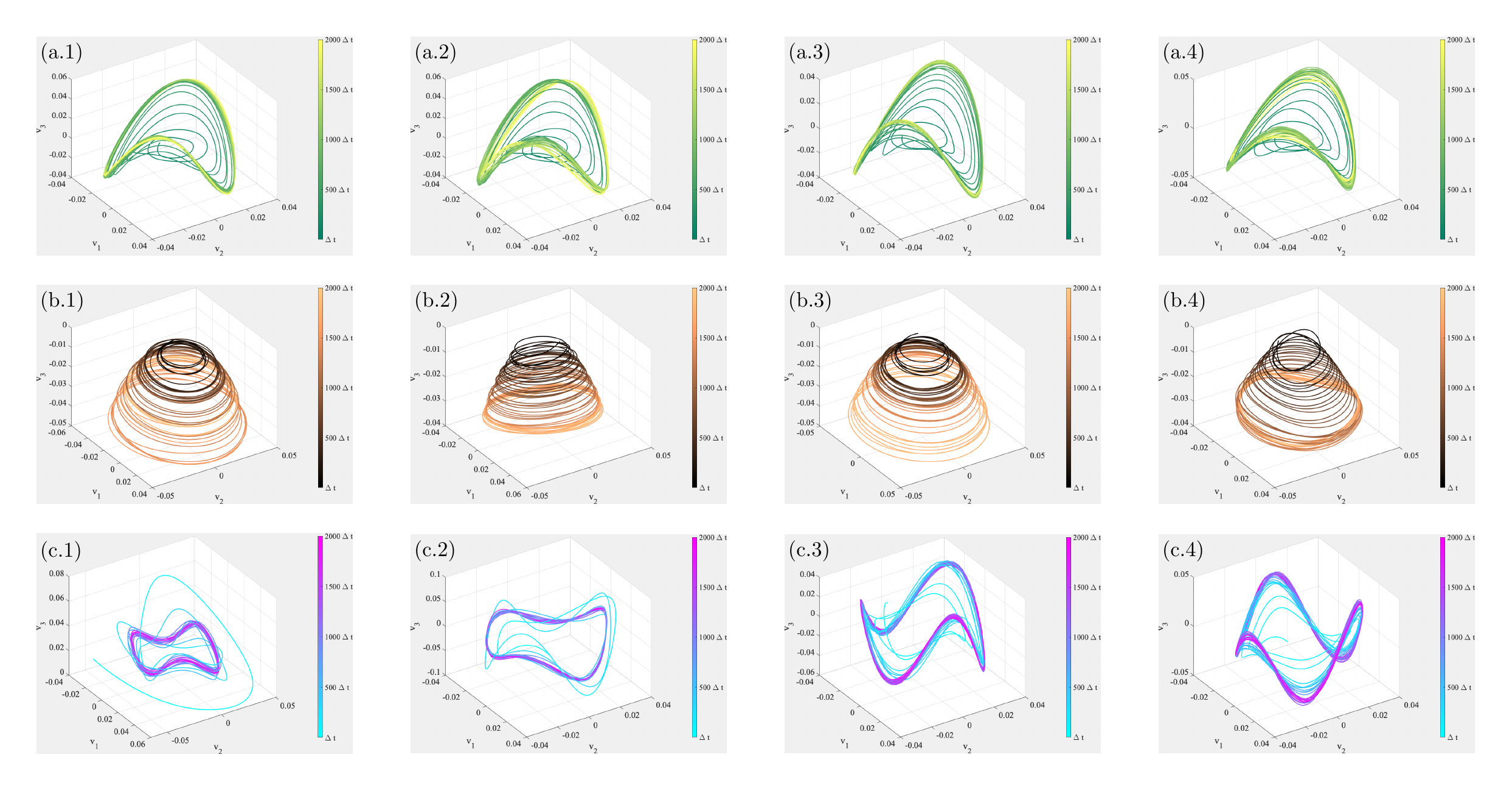}
    \caption{Embedding attractors of different clusters.} 
    \label{fig: f8}
\end{figure}

Based on the selected delay step $\tau$ and embedding dimension $d$, we can assemble the input matrix $\boldsymbol{X}$ for sparse learning (See \cref{eq: sec2_5}). Here, the dynamics of the embedded state variable at each time instance is plotted in three-dimensional space where singular vectors $V_1$, $V_2$, and $V_3$ that pack most of the energy contained in the embedded expression $\boldsymbol{X}$ are selected as the three major axes (See \cref{eq: sec2_6}). The selected sample trajectories from the first cluster have a horse saddle structure (See \cref{fig: f8}). The resulting attractor is skew-symmetric and diffuses from the inside out. When $t > 1000 \Delta t$, the trajectory asymptotically converges to the saddle contour. The reader is referred to the web version of this paper to better understand the convergence (Green states to yellow states). Meanwhile, samples from the second cluster have a truncated spinning top structure. Initialize a point at its apex upon which it is spun. Then, this point spins spirally from top to bottom and gradually expands the boundaries. It can be seen from \cref{fig: f8} that the attractor moves from a dark brown region to a Brownish orange region. The third cluster yields a squeezed sponge structure. In the beginning, the attractor chaotically oscillates in space. Later, it reaches a periodic orbit and traces ovals like a car on a hyper track (See \cref{fig: f8}).

\subsection{Part 4: sparse learning results}
\label{sec34}
\begin{figure}[H]
    \centering
    \includegraphics[width=0.85\textwidth]{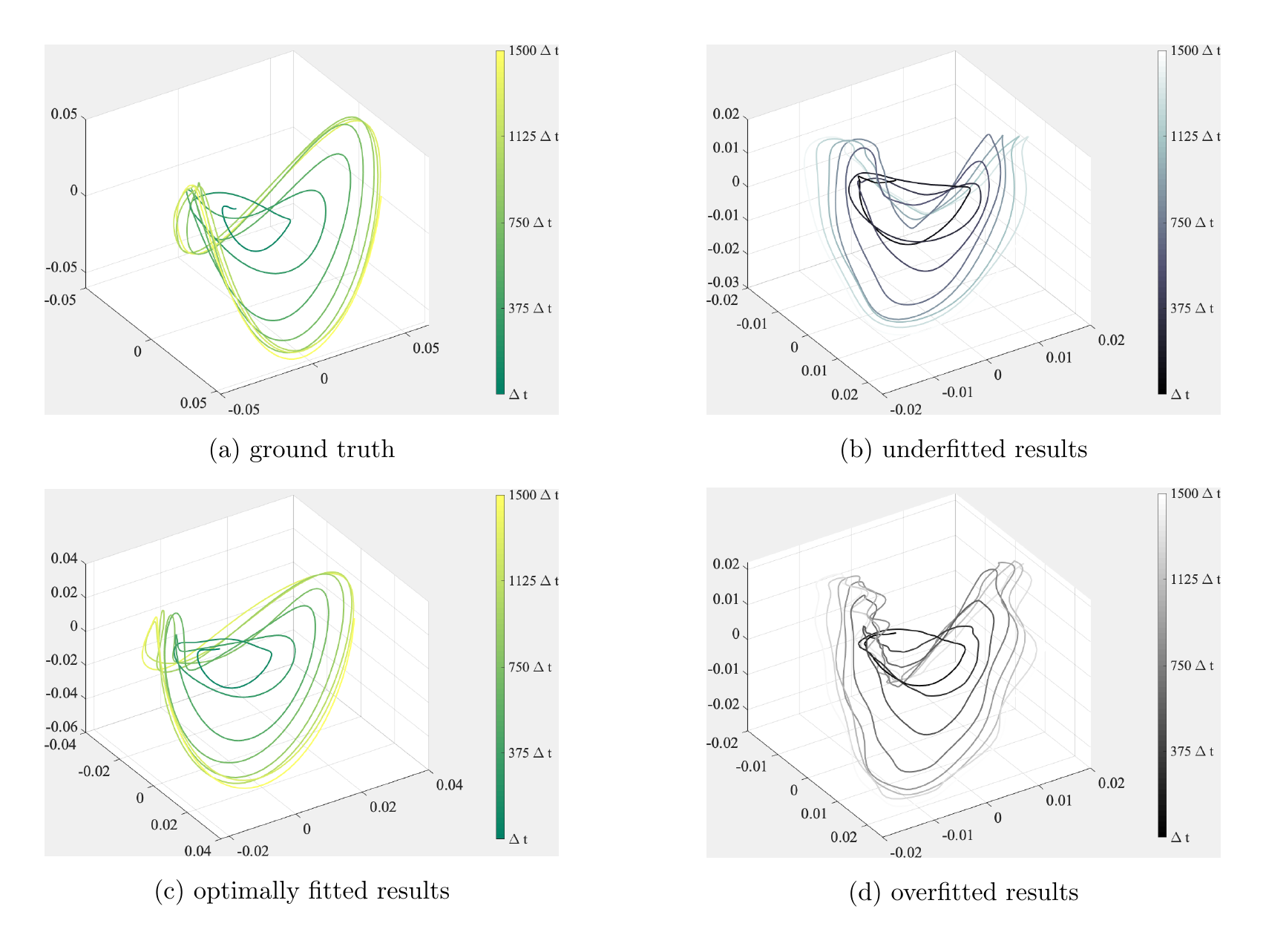}
    \caption{Reconstructed embedded attractor results from the third cluster.} 
    \label{fig: f9}
\end{figure}

The attractor is first reconstructed using the decomposed eigen components. We plot the data in the reconstructed phase space with the most energetic components $V_1, V_2$, and $V_3$ as $x(t), y(t)$, and $ z(t)$ for the first $1500$ time instances, as shown in \cref{fig: f9}. The remaining $500$ time instances ranging from $1500 \Delta t$ to $2000 \Delta t$ are used to assess prediction performance. Shown in \cref{fig: f9}, the optimally learned reconstructed attractor is qualitatively similar to the ground truth, and the trajectories of the reconstructed attractors are clearly smoother and less chaotic when fewer components are used. In practice, a large number of components will suffice to provide the necessary information for the reconstruction, and the least-squares estimates provided by \cref{eq: sec2_10} are unbiased. The variances of predicted values obtained by the fitted regression model, on the other hand, increase. As a result, the overfitting problem arises. For example, the number of components used in sparse learning for \cref{fig: f9} (b), (c), and (d) are $6$, $12$, and $22$, respectively. To select an appropriate number of components for sparse learning, one can use hard threshold learning methods such as asymptotic mean squared error-based singular value thresholding \cite{donoho2013optimal} and optical flow-based denoising \cite{abarbanel1993analysis} in addition to trial-and-error learning.

\begin{figure}[H]
    \centering
    \includegraphics[width=0.93\textwidth]{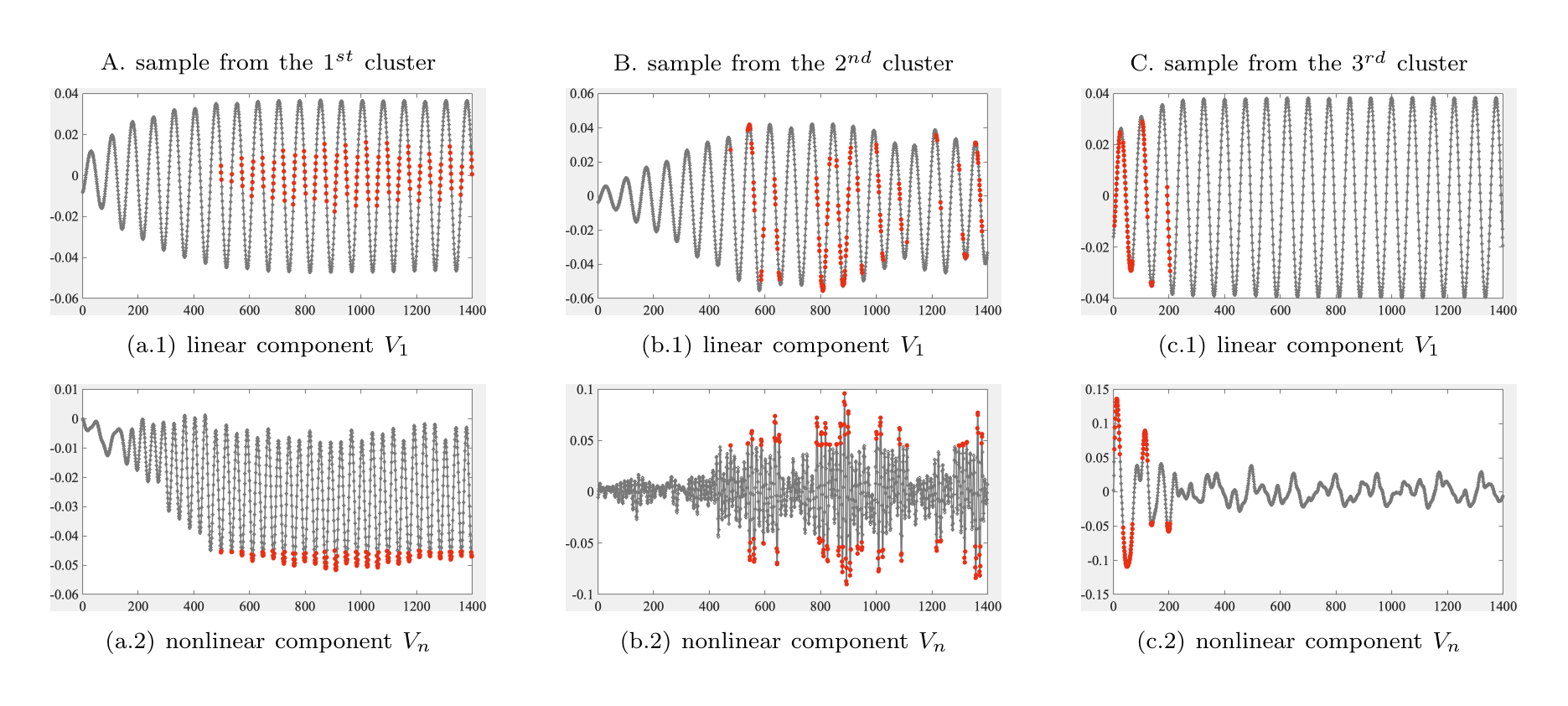}
    \caption{The linear and nonlinear (forcing) components of the sparse regression model.} 
    \label{fig: f10}
\end{figure}

\cref{fig: f10} shows the first linear component, $V_1$, as well as the last nonlinear component, $V_n$, of the SVD results of the embedded sloshing data $\boldsymbol{X}$, to investigate the functional roles played by linear and nonlinear components in the process of time-series analysis and attractor reconstruction. According to the computed results from sparse learning stated in the previous section, $n$ is 12, 14, and 7 for the sample from the first, second, and third clusters, respectively. The dynamics observed in the nonlinear component, $V_n$, is associated with intermittent bursts, lobe switching, and other chaotic points on the embedded attractor in many applications \cite{brunton2017chaos, khodkar2021data}. A threshold value of $\epsilon = 0.045$ is defined to identify the locations where the nonlinear component is active. When $V_n$ values exceed $epsilon$, the corresponding regions are isolated, classified as forcing active, and denoted by red dots.

\begin{figure}[b!]
    \centering
    \includegraphics[width=0.93\textwidth]{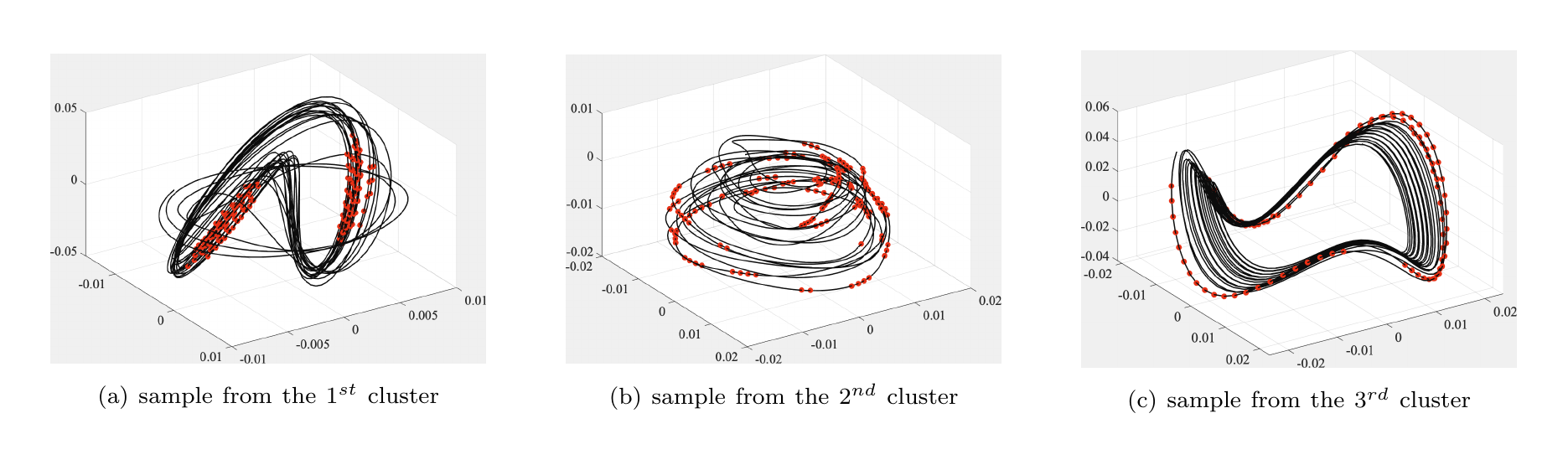}
    \caption{The embedded attractor with forcing active area.} 
    \label{fig: f11}
\end{figure}

As previously stated in \cite{abarbanel1993analysis, platt1993off}, the presence of on-off intermittency in an unstable reconstructed attractor can be explicitly revealed in a tuple of transformed coordinates. It is difficult to directly visualize the bursting/switching behavior in In \cref{fig: f10} using a single component, $V_1$. To help understand the on-off chaos, \cref{fig: f11} shows the underlying geometry of the calculated trajectories in three dimensions, with the forcing active regions highlighted. The red points in \cref{fig: f11} (a) indicate that lobe switching is about to happen. These are two groups of intermittent points, one in the basin of the left lobe and the other in the basin of the right lobe. This switching behavior is consistent with the complex wave motions observed in a partially filled tank, where the free-surface elevation first rises and then falls. The traveling waves are pushed in the opposite direction when they hit the vertical slat screens or the tank's sidewalls. The resulting nonlinear hydrodynamics causes the lobe switching on the embedded attractor to occur. Meanwhile, as illustrated in \cref{fig: f9}. (a), the initial states are more chaotic and less constrained by the embedded attractor. The evolution of the predicted trajectory reflects the same developmental trend. The highlighted locations in \cref{fig: f11}. (c) show strong nonlinearities at first, and these nonlinear effects subside when the skeleton of the attractor is formed.

\begin{table}[b!]
    \centering
    \begin{tabular}{l l l l}
    \hline
    Model & Fitted parameters & Result & $p-$value \\
    \hline
    Normal & $\mu = 0.0218, \sigma = 0.0073 $ & Rejected & $6.60 \times 10^{-22}$  \\
    Beta & $a = 10.1317, b = 455.1730 $ & Rejected & $1.60 \times 10^{-11}$  \\
    Exponential & $\mu = 0.0218$ & Rejected & $9.41 \times 10^{-223}$  \\
    EV & $\mu = 0.0258, \sigma = 0.0091$ & Rejected & $6.70 \times 10^{-47}$  \\
    Gamma & $a = 10.3905, b = 0.0021$ & Rejected & $2.75 \times 10^{-11}$  \\
    GEV & $k = 0.170, \mu = 0.004, \sigma = 0.018$ & Pass & $0.0961$  \\
    Lognormal & $\mu = -3.8761, \sigma = 0.3037$ & Rejected & $7.58 \times 10^{-7}$  \\
    Student t & $\mu = 0.019, \sigma = 0.004, v = 2.827$ & Rejected & $8.54 \times 10^{-13}$  \\
    Weibull & $A = 0.0243, B = 3.0241$ & Rejected & $1.20 \times 10^{-17}$  \\
    \hline
    \end{tabular}
    \begin{tablenotes}
    \footnotesize
    \item EV denotes extreme value distribution.
    \item GEV denotes generalized extreme value distribution.
    \end{tablenotes}
    \caption{Kolmogorov–Smirnov results of the selected sample from the first cluster.}
    \label{table: KSR_t1}
\end{table}

The statistics of the nonlinear component $V_n$ is essential to characterize the aforestated instability and intermittent phenomena. In practice, rare/extreme events of climate and ocean waves tend to have Non-Gaussian statistics \cite{takens1981detecting, broomhead1986extracting}. Based on this property, a set of hypothetic probability models are considered to describe the probability distribution of $V_n$, and maximum likelihood estimation (MLE) is applied to determine values for the fitted model's parameters. The one-sample Kolmogorov-Smirnov (K-S) test is performed to examine the goodness of fit and deliver the best hypothetic distribution model \cite{massey1951kolmogorov}. Specifically, the K-S test rejects the null hypothesis when the computed p-value is smaller than a predefined significance level, that is, $5 \%$ in this study. The test results for the sample selected from the first, second, and third cluster are outlined in \cref{table: KSR_t1}, \cref{table: KSR_t2}, and \cref{table: KSR_t3}, respectively. The table reports optimally fitted model parameters via MLE and decisions taken by comparing the test statistic with the critical value. Overall, the K-S test results indicate that the nonlinear component $V_n$ does not follow a normal distribution. The computed p-value is small, implying a doubt on the hypothetical probability distribution model's validity. \cref{table: KSR_t1} shows a generalized extreme value (GEV) distribution developed within extreme value theory provides the best performance in terms of characterizing the $1^{st}$ cluster's sample. Meanwhile, Student's t-distribution is demonstrated to the best statistical model to describe the other two samples (See \cref{table: KSR_t2} and \cref{table: KSR_t3}).

\begin{table}[h]
    \centering
    \begin{tabular}{l l l l}
    \hline
    Model & Fitted parameters & Result & $p-$value \\
    \hline
    Normal & $\mu = 0.0995, \sigma = 0.0266 $ & Rejected & $3.01 \times 10^{-5}$  \\
    Beta & $a = 10.8973, b = 98.6711 $ & Rejected & $3.61 \times 10^{-8}$  \\
    Exponential & $\mu = 0.0995$ & Rejected & $1.38 \times 10^{-210}$  \\
    EV & $\mu = 0.1135, \sigma = 0.0340$ & Rejected & $3.70 \times 10^{-28}$  \\
    Gamma & $a = 11.9046, b = 0.0084$ & Rejected & $1.04 \times 10^{-8}$  \\
    GEV & $k = -0.158, \mu = 0.026, \sigma = 0.088$ & Rejected & $1.90 \times 10^{-8}$  \\
    Lognormal & $\mu = -2.3500, \sigma = 0.3203$ & Rejected & $7.15 \times 10^{-15}$  \\
    Student t & $\mu = 0.098, \sigma = 0.018, v = 3.765$ & Pass & $0.9366$  \\
    Weibull & $A = 0.1093, B = 3.7700$ & Rejected & $2.05 \times 10^{-9}$  \\
    \hline
    \end{tabular}
    \caption{Kolmogorov–Smirnov results of the selected sample from the second cluster.}
    \label{table: KSR_t2}
\end{table}

\begin{table}[h]
    \centering
    \begin{tabular}{l l l l}
    \hline
    Model & Fitted parameters & Result & $p-$value \\
    \hline
    Normal & $\mu = 0.1197, \sigma = 0.0267 $ & Rejected & $3.41 \times 10^{-20}$  \\
    Beta & $a = 12.8022, b = 94.2699 $ & Rejected & $1.19 \times 10^{-38}$  \\
    Exponential & $\mu = 0.1197$ & Rejected & $1.04 \times 10^{-286}$  \\
    EV & $\mu = 0.1338, \sigma = 0.0369$ & Rejected & $4.75 \times 10^{-68}$  \\
    Gamma & $a = 13.9992, b = 0.0086$ & Rejected & $7.02 \times 10^{-41}$  \\
    GEV & $k = -0.178, \mu = 0.029, \sigma = 0.109$ & Rejected & $2.23 \times 10^{-30}$  \\
    Lognormal & $\mu = -2.1588, \sigma = 0.3184$ & Rejected & $1.09 \times 10^{-60}$  \\
    Student t & $\mu = 0.119, \sigma = 0.013, v = 2.377$ & Pass & $0.0504$  \\
    Weibull & $A = 0.1296, B = 4.2587$ & Rejected & $2.65 \times 10^{-40}$  \\
    \hline
    \end{tabular}
    \caption{Kolmogorov–Smirnov results of the selected sample from the third cluster.}
    \label{table: KSR_t3}
\end{table}

\begin{figure}[H]
    \centering
    \includegraphics[width=0.9\textwidth]{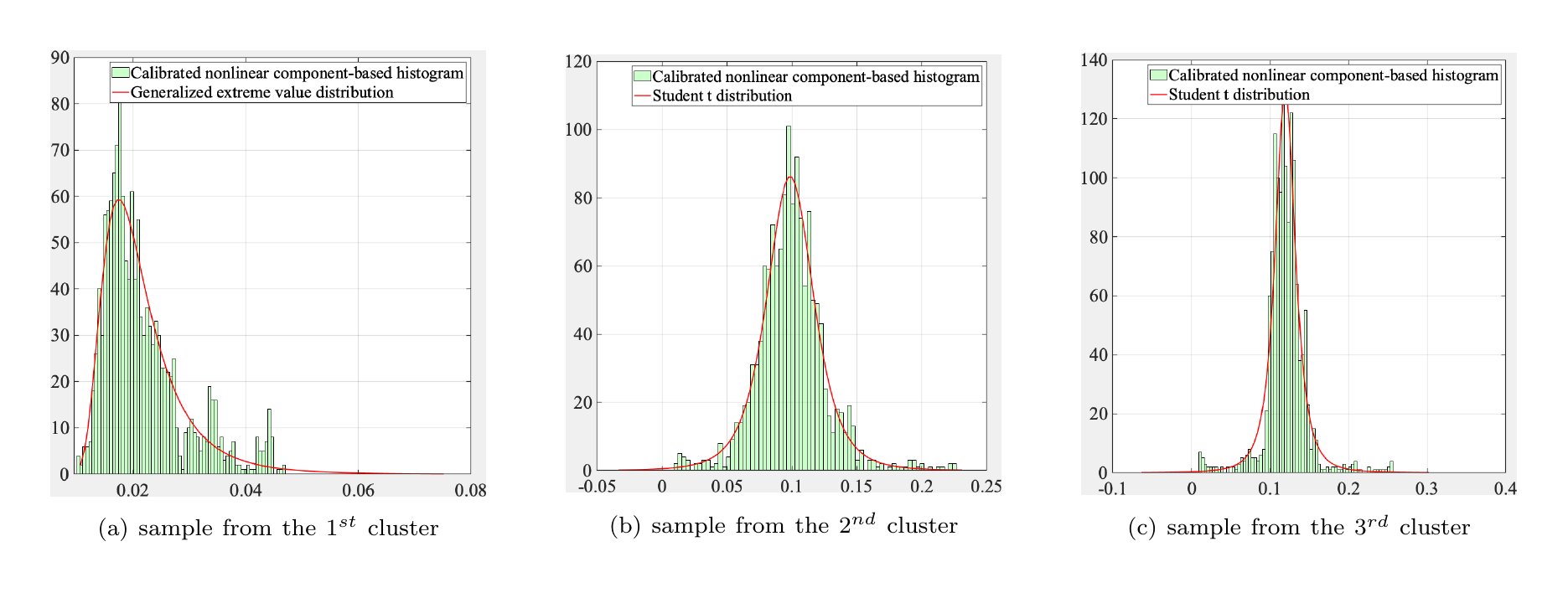}
    \caption{Optimally fitted probability density function (PDF) of the forcing term.} 
    \label{fig: f12}
\end{figure}

\cref{fig: f12} illustrates the histograms obtained from the calibrated $V_n$ and the probability density function of the best hypothetical probabilistic model identified by the K-S test. The length of the decomposed eigenvector $V_n$ is 1500, and the values have been shifted to the positive region to ensure the MLE algorithm's numerical stability and performance. It is realized by adding the absolute value of the minimum of $V_n$ to all the elements of $V_n$. As expected, the Non-Gaussian statistics are displayed in the fitted model's long tails compared with the normal distribution.

\begin{figure}[H]
    \centering
    \includegraphics[width=0.83\textwidth]{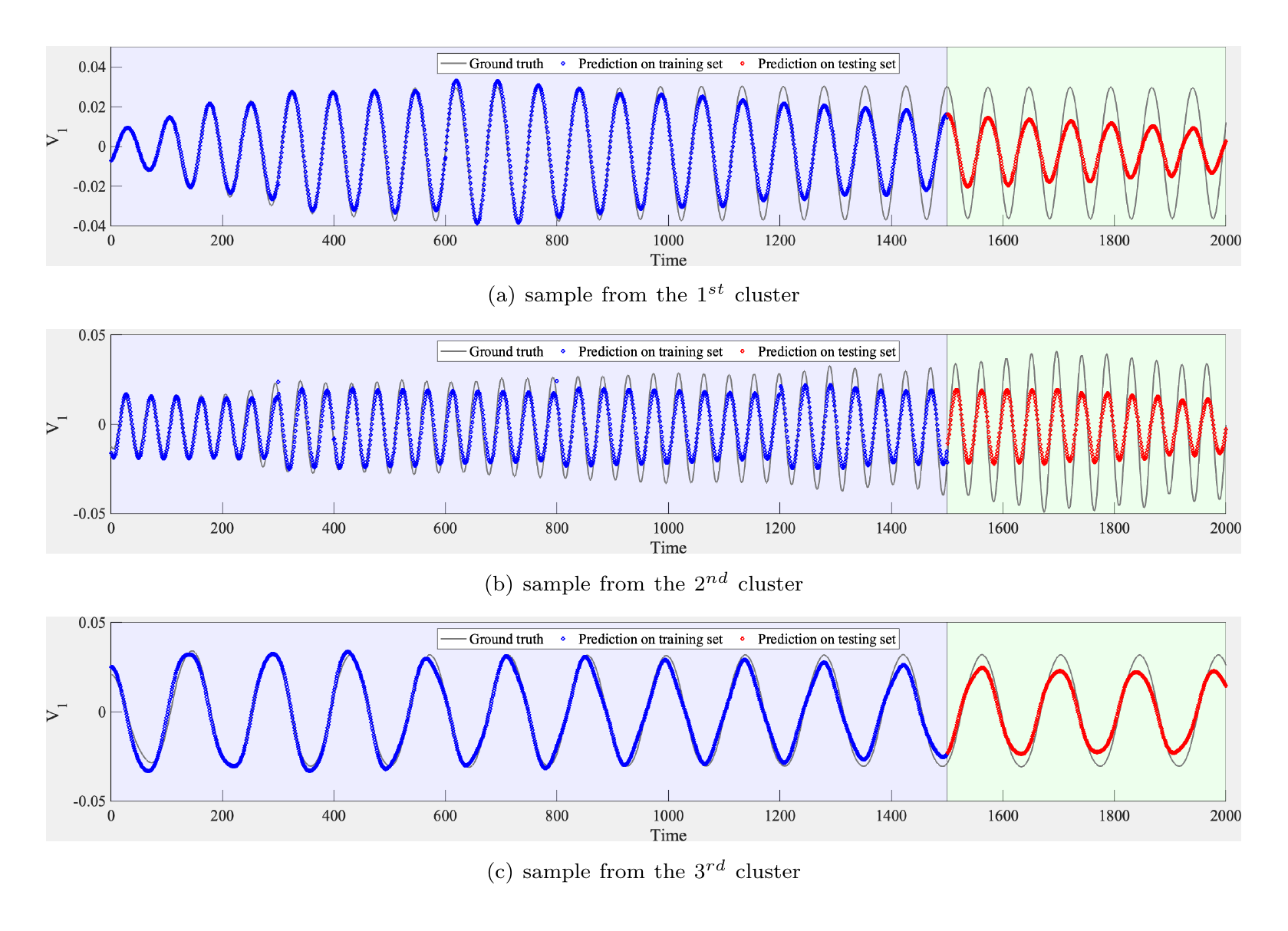}
    \caption{Validation of prediction performance on both training and testing sets.} 
    \label{fig: f13}
\end{figure}

It is important to validate the model's generalization ability once it is completely trained by examining the learned model's performance on data points not used in the training phase. Therefore, a test dataset is created to provide an unbiased evaluation of a model fit on the training dataset. In particular, an embedded sloshing sequence containing $2000$ time instances is split into a training set $V_1(\Delta t) \rightarrow V_1(1500 \Delta t)$ and a test set $V_1(1501 \Delta t) \rightarrow V_1(2000 \Delta t)$. The performance of using a well-trained model to predict the multi-steps of selected samples is given in \cref{fig: f13}. It can be seen that the model accurately captures the evolution pattern and envelope structure attached to the ground truth. For all samples selected, the difference between prediction and data gradually increases as time propagates.

\begin{figure}[H]
    \centering
    \includegraphics[width=0.85\textwidth]{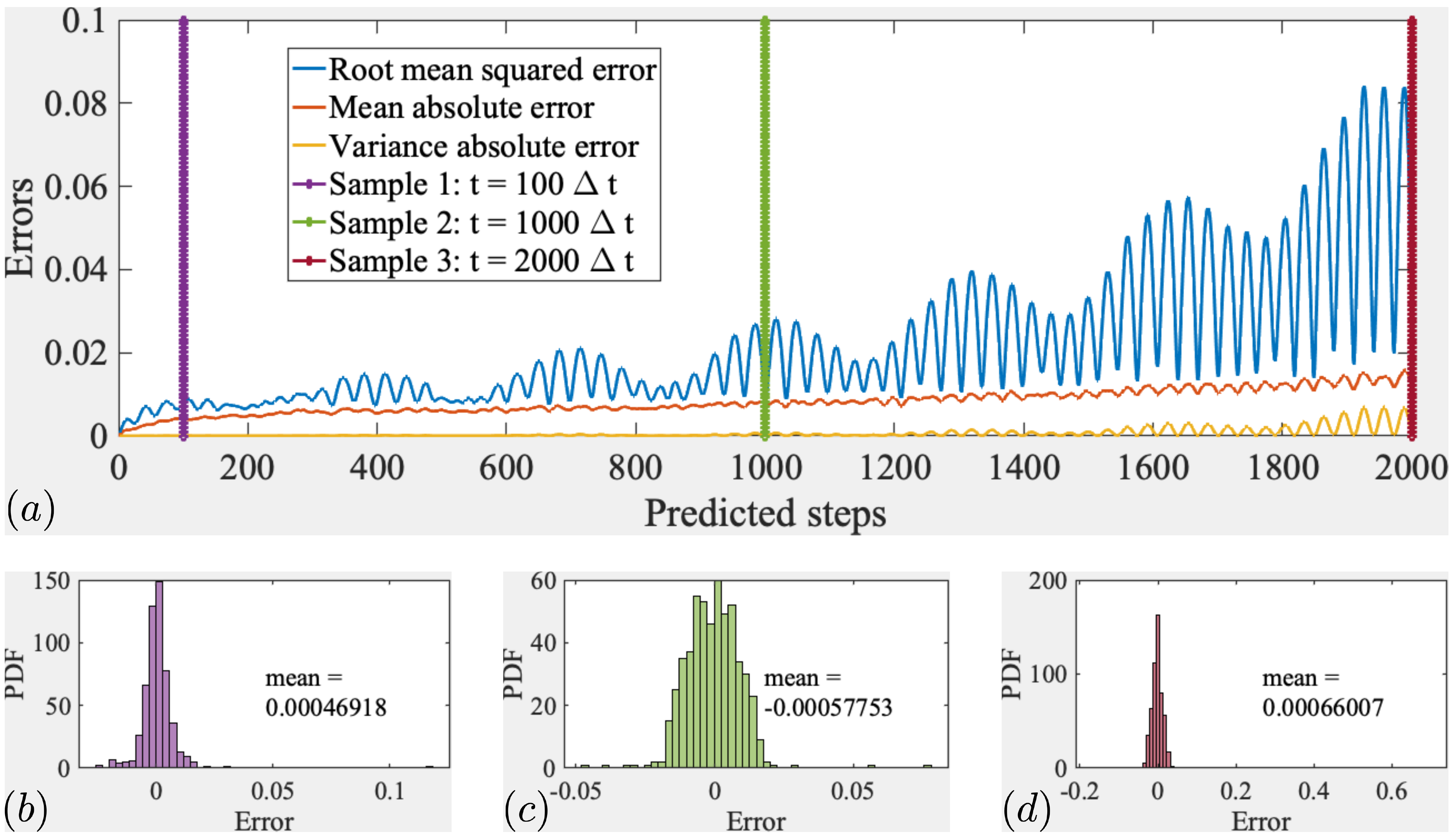}
    \caption{Validation of prediction performance on both training and testing sets.} 
    \label{fig: f14}
\end{figure}

Furthermore, we evaluate the accuracy of the prediction results using root mean square error (RMSE), mean absolute error (MAE), and variance absolute error (VAE). To statistically present the estimation errors of experiments with different configurations, all $539$ samples, each of which contains $2000$ time instances, are used to compute the evolution of the error distribution. \cref{fig: f14}. (a) presents the trend of the predefined error criteria, and \cref{fig: f14}. (b), (c), and (d) show the histograms of different errors at three selected time instances, namely, $100 \Delta t$, $1000 \Delta t$, and $2000 \Delta t$. Because VAE emphasizes large differences and RMSE/MAE is more used as a measure of dispersion, the evolution of VAE oscillates less rapidly than RMSE and MAE. It is observed that the mean error of those three histograms indicate the prediction error increases slightly as time propagates. The histograms are more concentrated near zero with a small center shift compared with an isotropic Gaussian distribution, which is often used to model errors in machine learning \cite{khodkar2021data, luo2020bayesian}.

\section{Concluding remarks}
\label{sec4}
This paper proposes a framework for characterizing the dynamics of nonlinear and non-Gaussian liquid sloshing. \textcolor{black}{In comparison to traditional system identification methods like wavelet, the proposed machine learning framework provides approximate linear representations of nonlinear dynamics. To address the data rank-deficiency, spectral decomposition is applied to time delay coordinates given a sequence of wave elevation. Then, using a set of decomposed eigenvectors, sequential sparse regression is utilized to identify an intermittently forced linear system. As a result, when compared to traditional approaches, the model's transparent data-driven structure and fast machine learning computation make it an excellent candidate for online feedback control tasks.} In future work, it is worthwhile to incorporate well-established controlling strategies into the current framework.

Through the use of a wide range of experimental data, we demonstrated the model is capable of accurately representing embedded sloshing sequences. Meanwhile, a series of statistical tests show that the model captures chaotic behaviors like bursting and switching in the embedded sequence. Various performance metrics are used to demonstrate the accuracy of the multi-step prediction results. \textcolor{black}{Admittedly, liquid sloshing can cause complex nonlinearities such as out-of-plane sloshing, spinning, irregular beat vibration, and pseudo-periodic motions. These nonlinearities are far more complex than the wave elevation dynamics we observed. Extending the methodology described here to different types of moving water waves, using either high fidelity CFD simulation data or more delicate experiment measurements, is a desirable first step toward developing a more robust and general data-driven simulation strategy for sloshing dynamics.}



\section*{Acknowledgments}
This work is supported by the U.S. Department of Energy (DOE), Office of Science, Advanced Scientific Computing Research under Award Number DE-SC-0012704.

\bibliographystyle{elsarticle-num}
\bibliography{sample}

\end{document}